\documentclass[]{interact}
\usepackage{url}
\usepackage{epstopdf}
\usepackage{subfigure}

\usepackage{natbib}
\bibpunct[, ]{(}{)}{;}{a}{}{,}
\renewcommand\bibfont{\fontsize{10}{12}\selectfont}

\theoremstyle{plain}

\theoremstyle{definition}

\theoremstyle{remark}

\begin{document}
\title{Towards Safe, Explainable, and Regulated Autonomous Driving}

\author{
\name{Shahin Atakishiyev\textsuperscript{a}\thanks{Corresponding author: Shahin Atakishiyev. Email: shahin.atakishiyev@ualberta.ca}, Mohammad Salameh\textsuperscript{b}, Hengshuai Yao\textsuperscript{a}, Randy Goebel\textsuperscript{a}}
\affil{\textsuperscript{a}Department of Computing Science, University of Alberta, Edmonton, Canada; \textsuperscript{b}Huawei Technologies Canada Co., Ltd, Edmonton, Canada}
}
\date{}
\maketitle

\begin{abstract}
There has been recent and growing interest in the development and deployment of autonomous vehicles, encouraged by the empirical successes of powerful artificial intelligence techniques (AI), especially in the applications of deep learning and reinforcement learning. However, as demonstrated by recent traffic accidents, autonomous driving technology is not fully reliable for safe deployment. As AI is the main technology behind the intelligent navigation systems of self-driving vehicles, both the stakeholders and transportation regulators require their AI-driven software architecture to be safe, explainable, and regulatory compliant. In this paper, we propose a design framework that integrates autonomous control, explainable AI (XAI), and regulatory compliance to address this issue, and then provide an initial validation of the framework with a critical analysis in a case study. Moreover, we describe relevant XAI approaches that can help achieve the goals of the framework.
\end{abstract}


\section{Introduction}

Autonomous driving is a rapidly growing field that has attracted increasing attention over the last decade. According to a recent report by Intel, the deployment of autonomous cars will reduce on-road travel by approximately 250 million hours and save about 585,000 lives per year between the years 2035 and 2045, just in the USA \citep{lanctot2017accelerating}. While these advantages certainly encourage the use of autonomous vehicles, there is also major public concern about the safety of this technology. This concern arises mainly from reports of recent accidents \citep{stanton2019models, yurtsever2020survey, board2020collision}  with the involvement of autonomous or semi-autonomous cars, primarily attributed to improper use of semi-autonomous functions. This issue is a major drawback, impeding self-operating vehicles from being acceptable by road users and society at a wider level.
As artificial intelligence techniques power autonomous vehicles' real-time decisions and actions, a malfunction of the vehicle’s intelligent control system is considered the main focus of analysis in such mishaps. Hence, both road users and regulators require that the AI systems of autonomous vehicles should be ``explainable,'' meaning that real-time decisions of such cars, particularly in critical traffic scenarios, should be intelligible in addition to being robust and safe. In this case

\begin{figure}[htp]
    \centering
    \includegraphics[width=8.8cm]{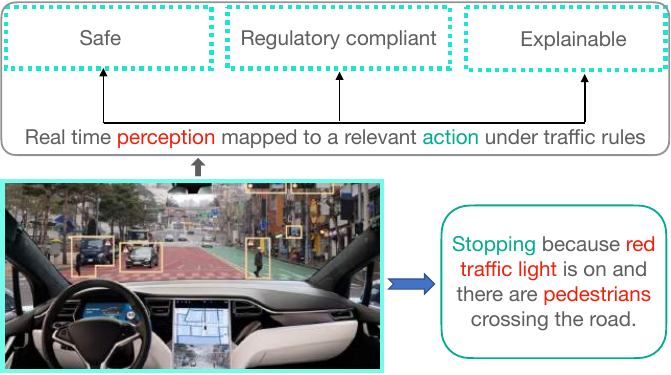}
    \caption{A graphical illustration of an autonomous car on its perception-action mapping that is safe, explainable, and regulatory compliant. \textit{Safe} because the vehicle drives under traffic rules and does not hit people and touch other objects. \textit{Explainable} because the vehicle provides a rationale for the taken action. \textit{Regulatory compliant} because the vehicle follows all the traffic rules and guidelines. The red-colored text implies perception and the green-colored text is the corresponding action. An image in the left corner from \citep{litman2021autonomous}.}
    \label{fig:end-to-end_control_framework}
\end{figure}
\hspace{-0.50cm}
explainability needs not only to provide transparency on an individual failure but also broadly inform the process of public transportation regulatory compliance \citep{compliance2020,dentons2021}. Urging the right to an explanation, several regulators have established safety and compliance standards for intelligent transportation systems. In this context, we are developing a general framework that can make autonomous vehicle manufacturers and involved users improve safety and regulatory compliance, and help autonomous vehicles become more publicly trustable and acceptable. With such a focus, our paper makes the following contributions:
\begin{itemize}
    \item We propose a general design framework for explainable autonomous driving systems and validate the framework with a use case;
    \item We present AI approaches from an algorithmic point of view that can help achieve explainability in autonomous driving and provide rationales behind an automated vehicle's real-time decisions.
\end{itemize}
The rest of the article is structured as follows. In Section 2, we provide a brief overview of modern autonomous driving and show the need for explainability in this technology. We introduce a relevant design framework in Section 3 and substantiate it with a case study in Section 4. Finally, in Section 5 we provide potential AI approaches that can help attain explainable autonomous driving systems and sum up the article with conclusions.

\section{A glance at state-of-the-art autonomous driving}
Autonomous cars, also known as self-driving or driverless cars, are capable of perceiving their environment and making real-time driving decisions with the help of intelligent driving control systems. To capture the operational environment, autonomous vehicles leverage a variety of passive sensors (e.g., collecting information from the surrounding without emitting a wave, such as visible spectrum cameras) and active sensors (e.g., sending a wave and receiving a reflected signal from objects, such as lidar). Sensor devices detect changes in the environment and enable the driving system of the car to make real-time decisions on the captured information \citep{campbell2018sensor, yeong2021sensor}. Current autonomous vehicles deployed on real roads are classified as having different levels of automation. SAE International has defined six different levels of autonomous driving based on the expected in-vehicle technologies and level of intelligent system, namely Level 0 - No automation,  Level 1 - Driving assistance, Level 2 - Partial automation, Level 3 - Conditional automation, Level 4 - High automation, and Level 5 - Full automation \citep{Shuttleworth}. 
The anticipated increase of automation levels escalates reliance on an intelligent driving system rather than a human driver, particularly in Level 3 and above. However, such vehicles, even at Level 3, have recently caused several road accidents, cited above, that have led to severe injuries or even loss of human lives. Why did the accident happen? What malfunction of the driving system led to the crash? 
These questions naturally raise serious ethical and safety issues and provide the motivation for \textit{explainable} AI (XAI) systems. In this context, the General Data Protection Regulation (GDPR) of the European Union (EU) established guidelines to promote a ``right to explanation'' for users, enacted in 2016 and carried into effect in May 2018 \citep{regulation2016}. 

In another example, The National Highway Traffic Safety Administration (NHTSA) of the US Department of Transportation has issued a federal guideline on automated vehicle policy to attain enhanced road safety \citep{national2016federal}. Current and future generation autonomous vehicles must comply with these emerging regulations, and their intelligent driving system should be explainable, transparent, and acceptably safe. In this regard, we propose a straightforward framework that considers the motivation for these requirements and then identify computational and legislative components that we believe are necessary for safe and transparent autonomous driving.

\section{An XAI framework for autonomous driving}

We present a general design framework in which methods for developing end-to-end autonomous driving, XAI, and regulatory compliance are connected. In this approach, the framework consists of three main components: (1) an end-to-end autonomous systems component, (2) a safety-regulatory compliance component, and (3) an XAI component. Explainability in the context of autonomous driving can be thought of as the ability of an intelligent driving system to provide transparency with comprehensible explanations  1) for any action, 2) to support failure investigation, and 3) in support of the process of public transportation regulatory compliance. We describe the role of the aforementioned components individually in the following subsections.

\subsection{An end-to-end autonomous systems component}

We need a simple but precise description of what we mean by ``end-to-end autonomous systems.'' To start, we need to be able to refer to the set of actions that {\it any} autonomous vehicle is capable of executing.  We consider the set of possible autonomous actions of automated vehicles as 
\[A = \{a_1, a_2, ... a_n\}.\]

Notice we consider the list of executable actions as a finite repertoire of actions that can be selected by a predictive model, whether that model is constructed by hand or by a machine learning algorithm or by some combination.  Example actions are things like ``turn right,'' or ``accelerate.''  For now, we refrain from considering complex actions like ``decelerate and turn right'' but acknowledge such actions will be possible in any given set $A$ of actions, depending on the vehicle.

We use the notation $C$ to denote an autonomous system controller and $E$ to denote the set of all possible autonomous system operating environments.  The overall function of an end-to-end controller is to map an environment $E$ to an action $A$; we have an informal description of the role of the autonomous system controller as
\[C : E \mapsto  A.\]

This mapping is intended to denote how {\it every} controller's responsibility is to map a perceived or sensed environment to an autonomous system function. We can provide a descriptive definition of an end-to-end autonomous controller as follows:
A control system $C$ is an {\it end-to-end control system} or {\it\bfseries eeC},
if $C$ is a total function that maps every instance of an environment 
 \[ e \in E\] to a relevant action  \[a \in A.\]
 

Even with such a high-level description, we can note that the most essential attribute of an {\it eeC} is that it provides a complete mapping from any sensed environment to an action selection.  We want this simple definition because we are not directly interested in sub-controllers for sub-actions; we are rather interested in autonomous controllers that provide complete control for high-level actions of any particular autonomous system, and which can be scrutinized for safety compliance for the whole scope of autonomous operation. Therefore, the {\it eeC} component is primarily responsible for perceiving the operational scene accurately using sensors such as video cameras, ultrasonics, lidar, radar, and other sensors, and enabling the car to take relevant actions.

\subsection{A safety-regulatory compliance component}
 The role of our framework's safety-regulatory component, {\it\bfseries srC}, is to represent the function of a regulatory agency,  whose primary role is to verify the safety of any configuration of $eeC$ for an autonomous vehicle's repertoire of actions \textit{A}.

We first note that safety compliance is a function that confirms the safety and reliability of an $eeC$ system. 
This could be as pragmatic as an inspection of individual vehicle safety (e.g., some jurisdictions require processes that certify essential safety functions of individual vehicles for re-licensing), but more and more is associated with sophisticated compliance testing of  $eeC$ components from manufacturers, to establish their public safety approval (e.g., \cite{Canada2021}).
The $srC$ components will vary widely from jurisdiction to jurisdiction (e.g., \cite{dentons2021}). In addition, AI and machine learning techniques have largely emphasized the construction of predictive models for $eeC$ systems. We can also note that all existing methods for software testing apply to this $srC$ task, and we expect the evolution of $srC$ processes will increasingly rely on the automation of compliance testing against all $eeC$ systems. The complexity of $srC$ systems lies within the scope of certified testing methods to confirm a threshold of safety. For example, from a compliance repertoire of $N$ safety tests, a regulator may require a 90\% performance of any particular $eeC$. With these expectations, the $srC$ 

\begin{figure*}[htp]
    \centering
    \includegraphics[width=0.9\textwidth]{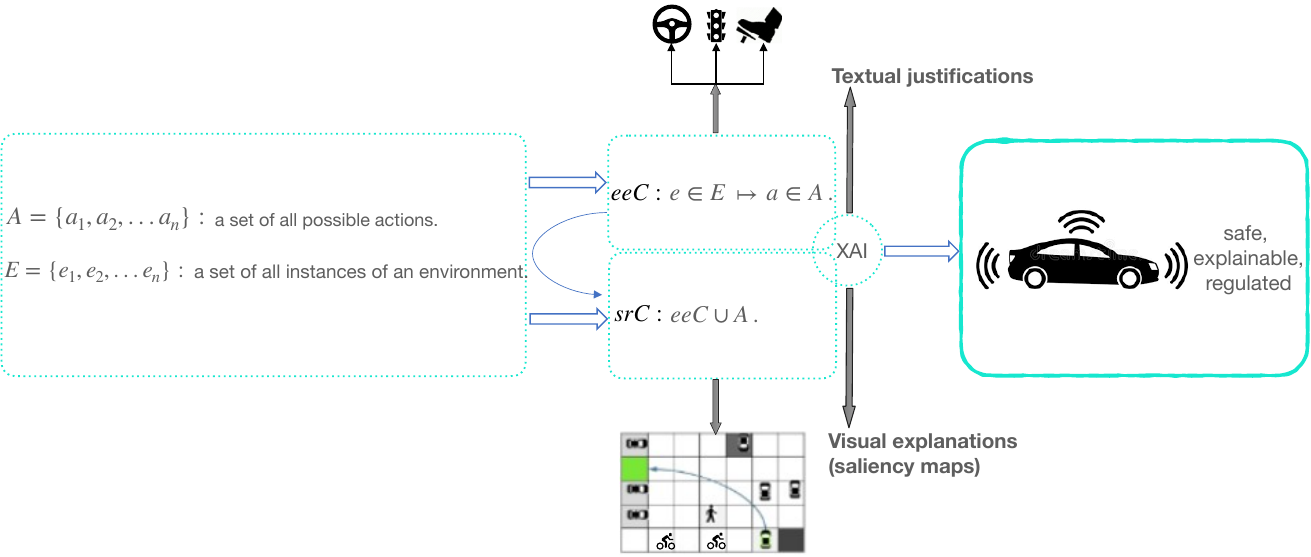}
    \caption{A graphical description of the proposed XAI framework for autonomous driving.}
    \label{fig:end-to-end_control_framework}
\end{figure*}\hspace{-0.6cm}
 component should guarantee the following safety concepts as defined by \citep{reschka2016safety}'s work: 
\begin{itemize}
    \item An autonomous car does not hit any static or dynamic objects on the road
    \item An autonomous car understands its performance abilities on the road
    \item An autonomous car does not block the traffic whether it is in motion or in a parking state
    \item An autonomous car does not exceed the speed limit specified in a road segment
    \item An autonomous car does not block the routes of an emergency vehicle 
    \item An autonomous car watches out for cars driving at a red traffic light 
    \item An autonomous car gets ready for a takeover by a backup driver in case automation capabilities are lost or malfunction
\end{itemize}
Thus, a proper combination of the $eeC$ and $srC$ components is a key to the correctness of the autonomous driving software system.

\subsection{An XAI component}
The third component of our framework is a general XAI mechanism. At the highest level, the XAI component of our framework is responsible for providing transparency on how an $eeC$ makes a selection from a repertoire of possible actions $A$. While the $eeC$ and $srC$ components have always been part of traditional autonomous driving systems, the explainability ability in real-time decisions remains unclear in the current state of the art. In particular, providing explanations in critical driving decisions is an essential factor to establish trust in the automation capability of autonomous driving technology. The difficulty of the challenge is directly related to those methods used to construct the control mechanism $C$ of the $eeC$. For example, a machine learning technique will need to address the selection of actions based on the interpretation of the sensed environment as accurately as possible.
This challenge is coupled with the demands created for the ``explainee,'' which may range from natural language interaction with a human driver (e.g., why did the system suggest having the driver take over steering?) to the potentially very complex setting of compliance safety parameters for regulatory compliance testing software. In terms of delivery, the explanations of the XAI component could be communicated in two ways as specified in Figure \ref{fig:end-to-end_control_framework}:
\begin{itemize}
    \item \textbf{Textual explanations:} Natural language-based explanations can justify decisions of an automated vehicle linguistically to the stakeholders. For example, if a vehicle changes its route at some time step unexpectedly, it could produce such an explanation: ``Changing predefined route \textit{because} there is a traffic gridlock ahead." 
    \item \textbf{Visual explanations:} In this case, explanations are delivered to end-users in a visual format. For instance, a passenger can see the visual perception of a car in an interface provided and judge appropriateness of real-time actions with respect to the visual information. Recently, \citep{greydanus2018visualizing}s' work, using saliency maps, shows how agents make their real-time decisions in Atari games. So, it would be worthwhile to apply the concept of saliency maps in the form of visual explanations to understand why a vehicle makes particular decisions in specific time steps.
\end{itemize}
The graphical description of the proposed framework is shown in Figure \ref{fig:end-to-end_control_framework}. As indicated, the three components of the framework are interrelated, and each component has a concrete role in the framework.  

\section{Case study: The role of the framework in a post-accident investigation}
Transportation regulators within their specific jurisdictions initiate a set of protocols to analyze the cause of traffic accidents with the involvement of autonomous vehicles. In such cases, the primarily investigated questions are: Who caused the accident? Is this a fault of a vehicle’s autonomous system, or did other road conditions and participants trigger the mishap? It is expected that autonomous cars can perform safe actions in all potential traffic scenarios that human drivers have commonly handled. ISO 26262 has established international safety standards for road automobiles that define safety rules and principles for passenger vehicles \citep{iso201126262}. This standard requires that a self-driving car provide intelligible and evidence-based rationales for the safety of its decisions in the operational environment.
Moreover, a vehicle should also provide information on potential residual risks of taking particular actions. As an example, we show a traffic scenario with an accident and describe the role of our framework to help identify the factors leading to this accident and to help further handle the situation appropriately according to traffic regulations. 

\begin{figure}[htp]
    \centering
    \includegraphics[width=8cm, height=4.3cm]{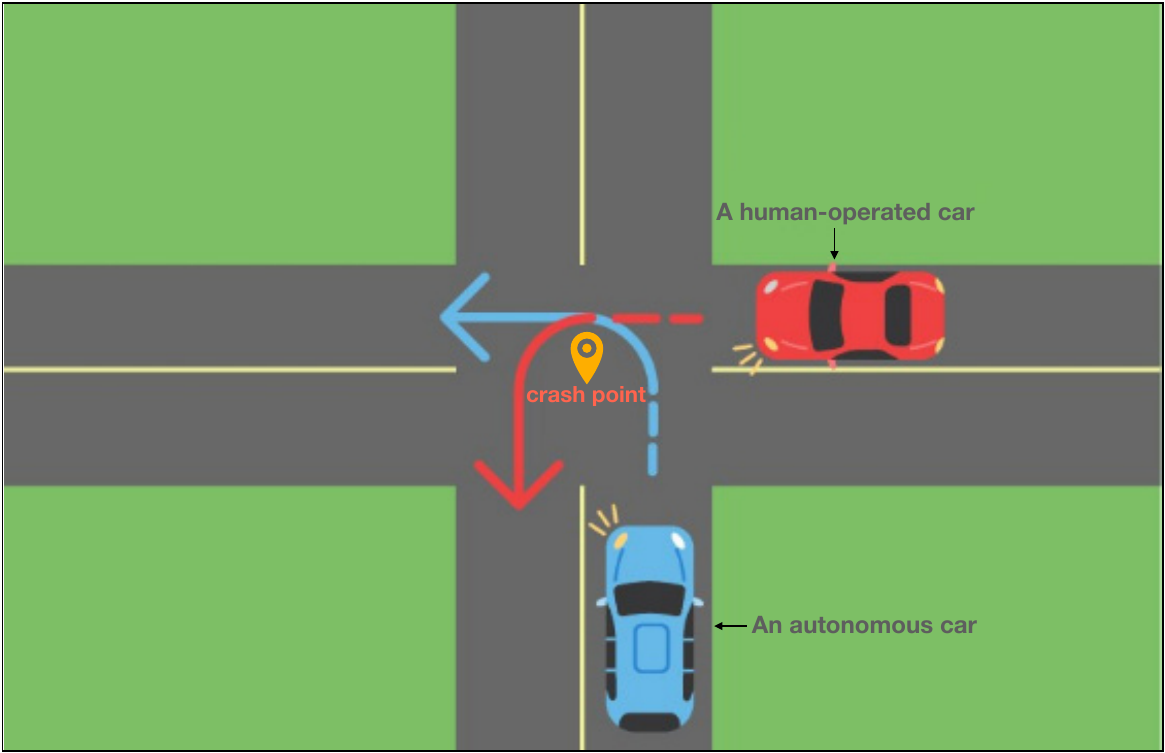}
    \caption{A traffic accident in an uncontrolled four-way intersection with autonomous and human-operated cars. Both vehicles try to turn left and incorrect coordination of time and maneuvers leads to a crash. As there may not be witnesses or other helpful arguments, an explainable AI-based driving system, storing a history of explanations alongside the relevant actions of an autonomous car can help understand which party made the wrong decision that caused the mishap. Graphics adapted from \citep{AMA_four_way}.}
    \label{fig:four_way_intersection}
\end{figure}

For our simple case study, we first assume a simple traffic scenario in an uncontrolled four-way intersection where an accident with the presence of an autonomous car (i.e., ego car) and another vehicle happens while both are attempting to make left turns in a given time step $t_n$ (Figure \ref{fig:four_way_intersection}). Accident investigators analyze the decisive actions of both cars at that particular time step. Suppose an autonomous vehicle records its action at $t_n$ and provides an \textit{explanation} as a justification for a taken action and the quantitative description of a residual risk associated with the performed action. Such functionality can help the investigators understand the main cause of the accident, i.e., whether an autonomous or other car took the wrong action. It immediately turns out that providing human interpretable justifications on a history of performed actions is a powerful tool both from a regulatory perspective in a post-accident analysis and from the lens of debugging and improving the existing automated system.\\
The other benefit of our framework is that it can help diminish the \textit{responsibility}, \textit{liability}, and \textit{semantic} gaps in autonomous driving, as identified by \citep{burton2020mind}. In terms of the responsibility gap, the framework, with its ability to explain a history of temporal actions, can help regulators and inspectors conclude whom to

\hspace{-0.55cm}
blame for the mishap. Once the leading actor of the accident is identified, financial responsibility can also be determined, and the liability gap can be resolved. Finally, the debuggability of intelligent driving can reduce the semantic gap: By inspecting the history of previously taken actions, AI engineers can take the opportunity to improve an existing system by deploying improved AI techniques. Therefore, we see that an intelligent driving system with explainability features and safety verification within the predefined standards has manifold benefits both from legal and consumer perspectives. 
\section{XAI hereafter: What to expect?}
We infer that end-to-end learning and motion trajectory of autonomous vehicles is based on precisely mapping perceived observations to corresponding actions. To date, perception has been mainly carried out through convolutional neural network (CNN) architectures and their augmented variations, and reinforcement learning (RL)-based approaches have proven computationally robust and adequate to map such real-time perception of states to relevant actions. Based on this intuition, we can note that explainable autonomous driving can be achieved by explaining the decisions of some combination of CNN/RL learning architecture. To conform to the framework presented, we provide directions for 1) explainable vision, 2) explainable RL, 3) representation of knowledge acquired from the operational environment, and 4) a question-driven software hierarchy for comprehensible action selection. 

\subsection{Use of trusted explainable vision: Understanding limitations and improving CNNs}
In recent studies, autonomous driving researchers and practitioners have already attempted to leverage CNN-based end-to-end learning. For example, Bojarski et al. \citep{bojarski2016end} used convolutional neural networks to map camera inputs to steering actions and obtained impressive results with minimal training data. Their CNN architecture learned a complete set of driving operations for driving in situations with and without lane markings in sunny, rainy, and cloudy weather conditions. In a further work, \citep{xu2017end} combined a dilated CNN with 
\begin{figure}[htp]
    \centering
    \includegraphics[scale=0.72]{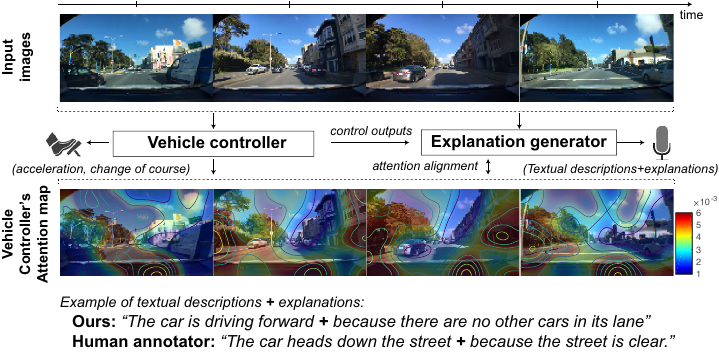}
    \caption{The intelligent driving system produces a textual explanation in a natural language, and a visual explanation with an attention mechanism on a taken action. Original source \citep{kim2018textual}.}
    \label{fig:controller-to-text}
\end{figure}
\hspace{-0.1cm}
a long-short term memory (LSTM) network to predict a self-driving vehicle's future motions. Another empirically successful end-to-end learning example from vision is \citep {toromanoff2020end}s' study, where the authors used reinforcement learning along with a convolutional encoder to drive an autonomous car safely while considering road users (e.g., pedestrians) and respecting driving rules (e.g., lane keeping, collision prevention, and traffic light detection). Their ablation study of the method proved effectiveness of the proposed approach by winning the ``Camera Only'' track of the CARLA competition challenge \citep{carlacompetition}. So, the end-to-end learning approach has demonstrated its effectiveness with an appropriate choice of real-time computational decision-making methods.  Consequently, explainable vision-directed actions for autonomous vehicles are based on how high-level features are used to detect and identify objects. CNNs, as the standard deep learning architectures, are ``black-box'' models, so there is a need to develop \textit{explainable CNN} (XCNN) architectures to comply with the requirements of the proposed framework. In this direction, there has been recent research on explaining predictions of CNNs: DeepLift \citep{shrikumar2017learning},  Class Activation Maps (CAM) \citep{zhou2016learning} and its extended versions such as Grad-CAM \citep{selvaraju2017grad}, Grad-CAM++ \citep{chattopadhay2018grad}, Guided Grad-CAM \citep{tang2019interpretable}, as well as heuristics-based Deep Visual Explanations \citep{babiker2017introduction, babiker2017using}. Motivated by such vision-based explainability methods, some recent studies have attempted to generate visual explanations for autonomous driving tasks.
In a related project, \citep{bojarski2016visualbackprop} developed a method, called VisualBackProp, that shows which set of pixels are primarily and the most influential in triggering the predictions of convolutional networks. \citep{kim2017interpretable} introduced a vision-based framework using \textit{causal attention} that shows what parts of an image control the steering wheel for appropriate actions. The authors extended their work in their further study and produce intelligible textual explanations on the vehicle’s actions \citep{kim2018textual}. Such \textit{post-hoc} explanations are a promising step towards transparent autonomous driving and can be helpful to understand critical decisions of an automated vehicle from legislative and stakeholders' perspectives, as mentioned in our case study.\\
Taking a further strategic step, in addition to these post-hoc explanations, it is strongly argued that autonomous driving control systems also need to provide \textit{intrinsic} explanations, where the developed system already becomes interpretable by design \citep{rudin2019stop}. While post-hoc explanations are useful for root cause identification in accident analysis, intrinsic explanations can be helpful to prevent such accidents. For instance, a back-up driver or an in-vehicle passenger can see textual explanations of critical decisions of a car with a suitable XAI interface while driving: in case these explanations

\begin{figure}[htp]
    \centering
    \includegraphics[width=8.8cm]{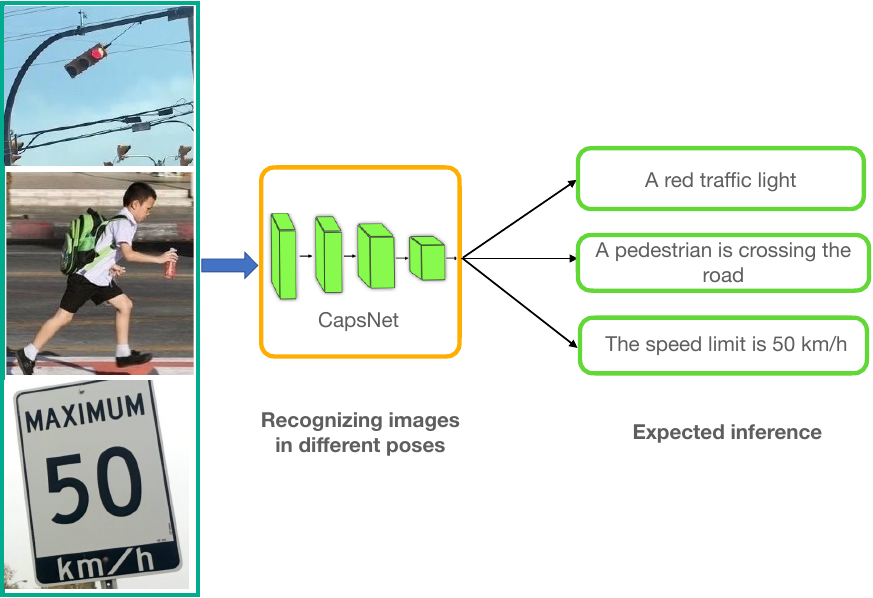}
    \caption{Three exemplary images in different poses: A traffic light sign affected by wind, a pedestrian leaning forward, and a speed limit sign with a changed vertical orientation. While CNNs work well with properly aligned images, changed shapes and poses may result in reduced accuracy in object recognition and classification tasks leading to further high-stakes consequences. CapsNet architectures, on the other hand, are able to detect objects with different poses. Similar to the provided images, we can see many other static and dynamic objects in everyday traffic that can be in various poses and shapes in specific times. CapsNet with its ability to detect objects in any poses can improve accuracy in object recognition and classification tasks in addition to its explainability advantage in perception tasks of autonomous driving.}
    \label{fig:CapsNet}
\end{figure}
\hspace{-0.5cm}
do not reflect the actual decisions, the back-up driver can control the car with their input (if available) or end the trip and prevent potential accidents in more complicated situations, foreseeable ahead. \\
There are also at least two additional limitations of CNNs. Firstly, they need an enormous volume of training data to perform well in image recognition and object classification tasks. Moreover, conventional CNNs require all possible orientations of a sample image during training, in order to accurately identify and recognize unseen images which have various orientations and poses. They do not learn pose-invariant objects and this property hampers CNN's modeling abilities. To overcome this downside of CNNs, \citep {hinton2011transforming} proposed the concept of \textit{capsules}, a set of artificial neurons that capture orientation and represent pose parameters such as size, position, and skewing of an object. In their subsequent work, \citep{NIPS2017_2cad8fa4} they used a \textit{dynamic routing} algorithm to train information passage between capsules at two successive layers. This feature enables the neural network to detect segments of an image and represent spatial representations differences between them although it is not yet clear how effective this can be. The hope is that the capsule neural network can detect an object in different shapes and poses even if a new sample of the same image has not previously been used in the training phase. Sabour et al.'s  most recent work \citep{sabour2021unsupervised} further improves the concept of dynamic routing and learns primary capsule encoders, which detect very tiny pieces of an image. \\
From the perspective of explainability, the learning mechanism of a capsule network makes this architecture intrinsically interpretable.  For example, in the medical domain, Sharoudnejad et al. have empirically shown that likelihood and instantiation parameter vector values provide rational explanations in classification tasks on the 28x28 MNIST dataset and MRI images of patients with brain cancer \citep{shahroudnejad2018improved}. Similarly, \citep{lalonde2020encoding} have shown that their explainable capsule network, called X-Caps, carries human-interpretable features in the vectors of its capsules encoded as high-level visual attributes.\\
Overall, both explainability and the ability to recognize objects in different poses make CapsNets promising in vision tasks of autonomous driving. In particular, static road objects such as speed signs and traffic lights often undergo the impact of adverse weather conditions (e.g., snow, wind) and collisions by vehicles that change their stance angle and form (e.g., see Figure \ref{fig:CapsNet}). As CNNs ultimately require regular shape and orientations of objects for real-time perception, CapsNet can improve their accuracy and consider the likelihood of the aforementioned pose, shape, and position changes along the traffic. Therefore, a CapsNet architecture's superior abilities with explainability and improved accuracy on object recognition tasks are promising research directions for the vision problems of autonomous driving.

\subsection{Explanation opportunity using model-based reinforcement learning}
Autonomous vehicles make sequential decisions throughout their motion trajectory within a setting formally characterized as a Markov Decision Process (MDP).  The field of reinforcement learning (e.g., \citep{sutton2018reinforcement}) provides an MDP implementation architecture whose high-level goal is to estimate the differential reward of any possible action and to incrementally adjust the priority of making decisions by computing a policy that produces ranking preferred actions. 
An RL agent’s interaction with the environment as an MDP can be implemented either as model-free or model-based RL. In a model-free setting (such as Q-learning), the algorithm does not have access to the dynamics of the environment (i.e., transition or reward function). It estimates the value function directly from a sensed experience \citep{sutton2018reinforcement}. So, model-free RL lacks explainability for learned policies. On the other hand, in the case of a model-based RL approach, an agent firstly tries to understand the world as prior knowledge, then develops a model to represent this world \citep{yao2012approximate, yao2014pseudo, sutton2018reinforcement, moerland2020model, kiran2021deep}. This approach in model-based RL is known as \textit{planning}. The idea of planning in RL is essential for understanding the explicit components of decision-making and has powered further model-based architectures (i.e., the Dyna architectures \citep {sutton1991dyna,linDyna, multi-stepDyna}). While in a model-free setting, an RL agent \textit{directly} learns a policy with the environment through interaction that produces a reward, in the Dyna and linear-Dyna style architectures an agent simultaneously learns a world model whilst learning an optimal policy through interactions with the world. Such a structure of planning makes it naturally \textit{explainable}. Whether based on approximations of state descriptions from the world or an imaginary state defined by the model, the planning process uses a model representation to generate a predicted future trajectory. According to the model projection, an ``optimal'' action is decided at each planning step, which provides a predicted state and a predicted reward. The predicted states and rewards can be analyzed and visualized for the planned trajectory, thus providing an explanation of why the agent favors the choice of a particular action at a specific time step. Within that context, the potential benefits and perspectives of using Dyna architectures and model-based RL are huge for XAI and autonomous driving. 

\subsection{Leveraging predictive knowledge}
It is important to specify how an agent could both represent and use the knowledge collected through interaction with the environment. This knowledge can be considered a compendium of predictions that the agent makes in anticipation of a selection from possible actions. Within the RL literature, such an approach to knowledge representation is called \textit{predictive knowledge}, and has received significant attention in RL studies \citep{drescher1991made, sutton2005temporal, sutton2011horde, white2015developing}.  An agent regularly expects a response from the environment by making many predictions about the dynamics of the environment in accordance with the autonomous system's behavior. Therefore, predictive knowledge, as one of the essential notions of reinforcement learning, can be considered as prior knowledge about the possible worlds in which the corresponding actions might be taken.
In order to consider a prediction as knowledge, a prediction should comply with the requirements of knowledge; first, a prediction should carry fundamental elements of epistemology - justification and truth, in itself \citep{kearney2019prediction}. Some, for example, \citep{white2015developing} and \citep{sherstan2020representation} have shown that General Value Functions (GVFs)\citep{sutton2011horde}, an architecture to learn numerous value functions from experience, are a promising proposal for the robust representation of predictive knowledge. In particular, \citep{sherstan2020representation} has empirically validated that GVFs, as a scalable representation, form a predictive model of an agent's interaction with its environment and can represent an operational environment. Recent works on GVFs have proven their value in perception \citep{graves2019perception} and policy learning problems of real-world autonomous driving \citep{graves2020learning}. Motivated by these results, it is beneficial to further investigate the concept of predictive knowledge and its representation with GVFs, to evaluate the robustness of such a formalism for autonomous driving. However, it is noteworthy to point out that GVFs commit to a particular encoding of predictive knowledge, which leaves their explanatory value undetermined.

\subsection{Temporal questions and question hierarchies}
If there is one common design principle that is shared by the autonomous driving industry, it is the commitment to arrange the control software in a hierarchical structure. 
To have a self-explainable decision procedure for autonomous driving, it is important that this hierarchy becomes {\em question driven}. In particular, the questions frequently arising while driving are those like ``Am I going to see the traffic lights turning yellow shortly?'', `` Why is the car in front braking?'', ``Will the car in the front right of me cut to my lane?'', etc. All these questions seem to come to us naturally when we drive. However, this concept of asking questions receives very little attention from current research on AI methods for autonomous driving. Understanding why human beings have this ability is vital to advancing the safety of autonomous machines. These questions don't emerge randomly but subconsciously; considering answers to these questions prepares us for a safe drive. 

To better understand the usefulness of the question-answering concept, assuming an autonomous car's intelligent driving system produces such a question-answer pair in a natural language in a particular time step: Q: ``\textit{Why} did the car change its lane?" - A: ``\textit{Because} an obstruction was observed ahead in the current lane." Such a question-answering pair would mainly be helpful in case of accident investigation, and also could help inspectors to ask further follow-up questions. \\
From the RL perspective, one potential approach to generate temporal questions based on the ongoing actions would be to use the concept of \textit{options} \citep{sutton1999between}. Options are the generalized concept of actions that has a policy for taking action with \textit{terminal} conditions. The \textit{option-critic architecture} was recently proposed by \citep{bacon2017option}.  Both internal policies and terminal conditions of options have been experimentally successful in end-to-end learning of options in the Arcade Learning Environment (ALE). 

The option-critic architecture is useful for temporal questions in autonomous driving; because driving-related questions are often temporal, new questions can be generated for the subsequent actions after just several seconds. Hence, it is important to study the formulation of questioning, with research focused on the generation, organization, and evaluation of questions for transparent autonomous driving \citep{zablocki2022explainability}. Note also that the sensitivity of driving decisions varies dynamically in real time, creating different levels of risk for the vehicle. In general, low-risk actions and options are preferred. However, in safety stringent situations, we need to explore efficiently (possibly in real-time) to manage possible dangers. This step requires a representation of the risks in decision making with a principled way of evaluating the risks to decide which risk level the vehicle is to undertake. Experiments by \citep{zhang2019quota} showed that considering no risks but only acting according to the maximum reward principle as in traditional RL is not always the best decision and can fail to discover the optimal policy. In contrast, acting according to a variety of levels of risk can find an optimal policy in environments with a diverse range of different transition and reward dynamics. We can thus infer and conclude that decision-making for autonomous vehicles is time-sensitive, often in sub-seconds, and a well-composed question hierarchy may not only help to select the present action but also determine subsequent actions that help sustain safe driving. The hierarchical structure can also provide temporal, informative, and reliable explanations in critical traffic situations with appropriate benefits.

\section{Conclusions}
As a result of our ongoing study, we presented a general design framework for XAI-based autonomous driving. We validated the framework with a case study of a post-accident analysis and showed that the proposed framework could have multi-faceted benefits to autonomous driving systems. First, the concept of the intrinsic and post-hoc explanations is not only limited to providing transparency on real-time decisions but also provides further opportunities to debug, fix, and enhance existing intelligent driving systems. Moreover, we showed that the principles of the proposed framework could address three actual issues, namely, responsibility, liability, and semantic gaps in the realm of autonomous driving. Finally, we presented some XAI approaches as future work and elucidated their potential in the explainability of autonomous driving.  \\
While the presented propositions are promising directions to follow, whether these concepts work effectively in a real driving environment remains unclear for now and is a limitation of our preliminary study. As a next step, we are performing empirical research conforming to the principles of the presented framework. Currently, we are trying to incorporate visual question answering with model-based reinforcement learning to achieve explainable vision mapped to explainable actions. Two interesting areas for deeper investigation are 1) a comparative analysis of model-free and model-based reinforcement learning approaches on the same task, and 2) whether the interpretability of an AI architecture for intelligent driving results in reduced accuracy of the interpretable algorithm compared to its original version on the same task. Therefore the contributions of this paper are mainly theoretical and further empirical studies are needed for proof of concept. We anticipate that further practical work based on these propositions can be helpful towards public acceptance of autonomous driving technology. 
  
\section{Acknowledgments}
We acknowledge support from the Alberta Machine Intelligence Institute
(Amii), from the Computing Science Department of the University of Alberta,
and the Natural Sciences and Engineering Research Council of Canada (NSERC). Shahin Atakishiyev also acknowledges support from the Ministry of Science and Education of the Republic of Azerbaijan.

\bibliographystyle{apalike}
\bibliography{main}

\begin{thebibliography}{}

\bibitem[Babiker and Goebel, 2017a]{babiker2017using}
Babiker, H. and Goebel, R. (2017a).
\newblock {Using KL-divergence to focus Deep Visual Explanation}.
\newblock {\em 31st Neural Information Processing Systems Conference (NIPS),
  Interpretable ML Symposium. Long Beach, CA, USA}.

\bibitem[Babiker and Goebel, 2017b]{babiker2017introduction}
Babiker, H. K.~B. and Goebel, R. (2017b).
\newblock {An Introduction to Deep Visual Explanation}.
\newblock {\em 31st Neural Information Processing Systems Conference (NIPS),
  Long Beach, CA, USA}.

\bibitem[Bacon et~al., 2017]{bacon2017option}
Bacon, P.-L., Harb, J., and Precup, D. (2017).
\newblock The option-critic architecture.
\newblock In {\em Proceedings of the AAAI Conference on Artificial
  Intelligence}, volume~31.

\bibitem[Bojarski et~al., 2016a]{bojarski2016visualbackprop}
Bojarski, M., Choromanska, A., Choromanski, K., Firner, B., Jackel, L., Muller,
  U., and Zieba, K. (2016a).
\newblock {Visualbackprop: visualizing CNNs for autonomous driving}.
\newblock {\em arXiv preprint arXiv:1611.05418}, 2.

\bibitem[Bojarski et~al., 2016b]{bojarski2016end}
Bojarski, M., Del~Testa, D., Dworakowski, D., Firner, B., Flepp, B., Goyal, P.,
  Jackel, L.~D., Monfort, M., Muller, U., Zhang, J., et~al. (2016b).
\newblock End to end learning for self-driving cars.
\newblock {\em arXiv preprint arXiv:1604.07316}.

\bibitem[Burton et~al., 2020]{burton2020mind}
Burton, S., Habli, I., Lawton, T., McDermid, J., Morgan, P., and Porter, Z.
  (2020).
\newblock Mind the gaps: Assuring the safety of autonomous systems from an
  engineering, ethical, and legal perspective.
\newblock {\em Artificial Intelligence}, 279:103201.

\bibitem[{{Cameron McKay}}, 2021]{AMA_four_way}
{{Cameron McKay}} (2021).
\newblock Uncontrolled intersection rules in {Alberta}.
\newblock
  \url{https://ama.ab.ca/articles/uncontrolled-intersection-rules-in-alberta}.
\newblock {Accessed on April 10, 2022}.

\bibitem[Campbell et~al., 2018]{campbell2018sensor}
Campbell, S., O'Mahony, N., Krpalcova, L., Riordan, D., Walsh, J., Murphy, A.,
  and Ryan, C. (2018).
\newblock Sensor technology in autonomous vehicles: A review.
\newblock In {\em 2018 29th Irish Signals and Systems Conference (ISSC)}, pages
  1--4. IEEE.

\bibitem[{CARLA's blog}, 2019]{carlacompetition}
{CARLA's blog} (2019).
\newblock {The CARLA Autonomous Driving Challenge}.
\newblock (Accessed on November 1, 2021).

\bibitem[Chattopadhay et~al., 2018]{chattopadhay2018grad}
Chattopadhay, A., Sarkar, A., Howlader, P., and Balasubramanian, V.~N. (2018).
\newblock Grad-cam++: Generalized gradient-based visual explanations for deep
  convolutional networks.
\newblock In {\em 2018 IEEE Winter Conference on Applications of Computer
  Vision (WACV)}, pages 839--847. IEEE.

\bibitem[Dentons, 2021]{dentons2021}
Dentons (2021).
\newblock {Global Guide to Autonomous Vehicles 2021}.

\bibitem[Drescher, 1991]{drescher1991made}
Drescher, G.~L. (1991).
\newblock {\em {Made-up minds: A constructivist approach to artificial
  intelligence}}.
\newblock MIT Press.

\bibitem[GDPR, 2016]{regulation2016}
GDPR (2016).
\newblock {Regulation EU 2016/679 of the European Parliament and of the Council
  of 27 April 2016}.
\newblock {\em Official Journal of the European Union}.

\bibitem[Graves et~al., 2020]{graves2020learning}
Graves, D., Nguyen, N.~M., Hassanzadeh, K., and Jin, J. (2020).
\newblock Learning predictive representations in autonomous driving to improve
  deep reinforcement learning.
\newblock {\em arXiv preprint arXiv:2006.15110}.

\bibitem[Graves et~al., 2019]{graves2019perception}
Graves, D., Rezaee, K., and Scheideman, S. (2019).
\newblock Perception as prediction using general value functions in autonomous
  driving applications.
\newblock In {\em 2019 IEEE/RSJ International Conference on Intelligent Robots
  and Systems (IROS)}, pages 1202--1209. IEEE.

\bibitem[Greydanus et~al., 2018]{greydanus2018visualizing}
Greydanus, S., Koul, A., Dodge, J., and Fern, A. (2018).
\newblock Visualizing and understanding atari agents.
\newblock In {\em International Conference on Machine Learning}, pages
  1792--1801. PMLR.

\bibitem[Hinton et~al., 2011]{hinton2011transforming}
Hinton, G.~E., Krizhevsky, A., and Wang, S.~D. (2011).
\newblock Transforming auto-encoders.
\newblock In {\em International Conference on Artificial Neural Networks},
  pages 44--51. Springer.

\bibitem[{ISO 26262}, 2011]{iso201126262}
{ISO 26262} (2011).
\newblock Road vehicles-functional safety.
\newblock {\em International Standard ISO/FDIS}, 26262.

\bibitem[Kearney and Pilarski, 2019]{kearney2019prediction}
Kearney, A. and Pilarski, P.~M. (2019).
\newblock When is a prediction knowledge?
\newblock {\em arXiv preprint arXiv:1904.09024}.

\bibitem[Kim and Canny, 2017]{kim2017interpretable}
Kim, J. and Canny, J. (2017).
\newblock Interpretable learning for self-driving cars by visualizing causal
  attention.
\newblock In {\em Proceedings of the IEEE International Conference on Computer
  Vision}, pages 2942--2950.

\bibitem[Kim et~al., 2018]{kim2018textual}
Kim, J., Rohrbach, A., Darrell, T., Canny, J., and Akata, Z. (2018).
\newblock Textual explanations for self-driving vehicles.
\newblock In {\em Proceedings of the European Conference on Computer Vision
  (ECCV)}, pages 563--578.

\bibitem[Kiran et~al., 2021]{kiran2021deep}
Kiran, B.~R., Sobh, I., Talpaert, V., Mannion, P., Al~Sallab, A.~A., Yogamani,
  S., and P{\'e}rez, P. (2021).
\newblock Deep reinforcement learning for autonomous driving: A survey.
\newblock {\em IEEE Transactions on Intelligent Transportation Systems}.

\bibitem[LaLonde et~al., 2020]{lalonde2020encoding}
LaLonde, R., Torigian, D., and Bagci, U. (2020).
\newblock Encoding visual attributes in capsules for explainable medical
  diagnoses.
\newblock In {\em International Conference on Medical Image Computing and
  Computer-Assisted Intervention}, pages 294--304. Springer.

\bibitem[Lanctot et~al., 2017]{lanctot2017accelerating}
Lanctot, R. et~al. (2017).
\newblock Accelerating the future: The economic impact of the emerging
  passenger economy.
\newblock {\em Strategy Analytics}, 5.

\bibitem[Litman, 2021]{litman2021autonomous}
Litman, T. (2021).
\newblock Autonomous vehicle implementation predictions: Implications for
  transport planning.

\bibitem[Moerland et~al., 2020]{moerland2020model}
Moerland, T.~M., Broekens, J., and Jonker, C.~M. (2020).
\newblock Model-based reinforcement learning: A survey.
\newblock {\em arXiv preprint arXiv:2006.16712}.

\bibitem[{NHTSA}, 2016]{national2016federal}
{NHTSA} (2016).
\newblock {\em Federal automated vehicles policy: Accelerating the next
  revolution in roadway safety}.
\newblock US Department of Transportation.

\bibitem[NTSB, 2020]{board2020collision}
NTSB (2020).
\newblock Collision between a sport utility vehicle operating with partial
  driving automation and a crash attenuator {Mountain View}, {California}.

\bibitem[Reschka, 2016]{reschka2016safety}
Reschka, A. (2016).
\newblock Safety concept for autonomous vehicles.
\newblock In {\em Autonomous Driving}, pages 473--496. Springer.

\bibitem[Rudin, 2019]{rudin2019stop}
Rudin, C. (2019).
\newblock Stop explaining black box machine learning models for high stakes
  decisions and use interpretable models instead.
\newblock {\em Nature Machine Intelligence}, 1(5):206--215.

\bibitem[Sabour et~al., 2017]{NIPS2017_2cad8fa4}
Sabour, S., Frosst, N., and Hinton, G.~E. (2017).
\newblock Dynamic routing between capsules.
\newblock In {\em Advances in Neural Information Processing Systems},
  volume~30.

\bibitem[Sabour et~al., 2021]{sabour2021unsupervised}
Sabour, S., Tagliasacchi, A., Yazdani, S., Hinton, G., and Fleet, D.~J. (2021).
\newblock Unsupervised part representation by flow capsules.
\newblock In {\em International Conference on Machine Learning}, pages
  9213--9223. PMLR.

\bibitem[Selvaraju et~al., 2017]{selvaraju2017grad}
Selvaraju, R.~R., Cogswell, M., Das, A., Vedantam, R., Parikh, D., and Batra,
  D. (2017).
\newblock {Grad-CAM: Visual explanations from deep networks via gradient-based
  localization}.
\newblock In {\em Proceedings of the IEEE International Conference on Computer
  Vision}, pages 618--626.

\bibitem[Shahroudnejad et~al., 2018]{shahroudnejad2018improved}
Shahroudnejad, A., Afshar, P., Plataniotis, K.~N., and Mohammadi, A. (2018).
\newblock Improved explainability of capsule networks: Relevance path by
  agreement.
\newblock In {\em 2018 IEEE Global Conference on Signal and Information
  Processing (GlobalSIP)}, pages 549--553. IEEE.

\bibitem[Sherstan, 2020]{sherstan2020representation}
Sherstan, C. (2020).
\newblock Representation and general value functions.
\newblock {\em PhD thesis, University of Alberta}.

\bibitem[Shrikumar et~al., 2017]{shrikumar2017learning}
Shrikumar, A., Greenside, P., and Kundaje, A. (2017).
\newblock Learning important features through propagating activation
  differences.
\newblock In {\em International Conference on Machine Learning}, pages
  3145--3153. PMLR.

\bibitem[Shuttleworth, 2019]{Shuttleworth}
Shuttleworth, J. (2019).
\newblock {SAE} standard news: J3016 automated-driving graphic update.
\newblock
  \url{https://www.sae.org/news/2019/01/sae-updates-j3016-automated-driving-graphic}.
\newblock {Accessed online} on August 16, 2021.

\bibitem[Slagter and Voster, 2017]{compliance2020}
Slagter, R. and Voster, R. (2017).
\newblock {Autonomous Compliance Standing on the shoulders of RegTech!}
\newblock {\em Compact}.

\bibitem[Stanton et~al., 2019]{stanton2019models}
Stanton, N.~A., Salmon, P.~M., Walker, G.~H., and Stanton, M. (2019).
\newblock {Models and methods for collision analysis: a comparison study based
  on the Uber collision with a pedestrian}.
\newblock {\em Safety Science}, 120:117--128.

\bibitem[Sutton, 1991]{sutton1991dyna}
Sutton, R.~S. (1991).
\newblock Dyna, an integrated architecture for learning, planning, and
  reacting.
\newblock {\em ACM Sigart Bulletin}, 2(4):160--163.

\bibitem[Sutton and Barto, 2018]{sutton2018reinforcement}
Sutton, R.~S. and Barto, A.~G. (2018).
\newblock {\em Reinforcement learning: An introduction}.
\newblock MIT Press, {Second} edition.

\bibitem[Sutton et~al., 2011]{sutton2011horde}
Sutton, R.~S., Modayil, J., Delp, M., Degris, T., Pilarski, P.~M., White, A.,
  and Precup, D. (2011).
\newblock Horde: A scalable real-time architecture for learning knowledge from
  unsupervised sensorimotor interaction.
\newblock In {\em The 10th International Conference on Autonomous Agents and
  Multiagent Systems-Volume 2}, pages 761--768.

\bibitem[Sutton et~al., 1999]{sutton1999between}
Sutton, R.~S., Precup, D., and Singh, S. (1999).
\newblock {Between MDPs and semi-MDPs: A framework for temporal abstraction in
  reinforcement learning}.
\newblock {\em Artificial Intelligence}, 112(1-2):181--211.

\bibitem[Sutton et~al., 2008]{linDyna}
Sutton, R.~S., Szepesv{\'{a}}ri, C., Geramifard, A., and Bowling, M.~H. (2008).
\newblock Dyna-style planning with linear function approximation and
  prioritized sweeping.
\newblock In McAllester, D.~A. and Myllym{\"{a}}ki, P., editors, {\em {UAI}
  2008, Proceedings of the 24th Conference in Uncertainty in Artificial
  Intelligence, Helsinki, Finland, July 9-12, 2008}, pages 528--536. {AUAI}
  Press.

\bibitem[Sutton and Tanner, 2005]{sutton2005temporal}
Sutton, R.~S. and Tanner, B. (2005).
\newblock Temporal-difference networks.
\newblock In {\em Advances in Neural Information Processing Systems}, pages
  1377--1384.

\bibitem[Tang et~al., 2019]{tang2019interpretable}
Tang, Z., Chuang, K.~V., DeCarli, C., Jin, L.-W., Beckett, L., Keiser, M.~J.,
  and Dugger, B.~N. (2019).
\newblock {Interpretable classification of Alzheimer’s disease pathologies
  with a convolutional neural network pipeline}.
\newblock {\em Nature Communications}, 10(1):1--14.

\bibitem[Toromanoff et~al., 2020]{toromanoff2020end}
Toromanoff, M., Wirbel, E., and Moutarde, F. (2020).
\newblock End-to-end model-free reinforcement learning for urban driving using
  implicit affordances.
\newblock In {\em Proceedings of the IEEE/CVF Conference on Computer Vision and
  Pattern Recognition}, pages 7153--7162.

\bibitem[{Transport Canada}, 2021]{Canada2021}
{Transport Canada} (2021).
\newblock {Guidelines for testing automated driving systems in Canada}.
\newblock {\em Ministry of Transportation of Canada}.

\bibitem[White, 2015]{white2015developing}
White, A. (2015).
\newblock Developing a predictive approach to knowledge.
\newblock {\em PhD thesis, University of Alberta}.

\bibitem[Xu et~al., 2017]{xu2017end}
Xu, H., Gao, Y., Yu, F., and Darrell, T. (2017).
\newblock End-to-end learning of driving models from large-scale video
  datasets.
\newblock In {\em Proceedings of the IEEE Conference on Computer Vision and
  Pattern Recognition}, pages 2174--2182.

\bibitem[Yao et~al., 2009]{multi-stepDyna}
Yao, H., Bhatnagar, S., Diao, D., Sutton, R.~S., and Szepesv\'{a}ri, C. (2009).
\newblock Multi-step dyna planning for policy evaluation and control.
\newblock In Bengio, Y., Schuurmans, D., Lafferty, J., Williams, C., and
  Culotta, A., editors, {\em Advances in Neural Information Processing
  Systems}, volume~22. Curran Associates, Inc.

\bibitem[Yao and Szepesv{\'a}ri, 2012]{yao2012approximate}
Yao, H. and Szepesv{\'a}ri, C. (2012).
\newblock Approximate policy iteration with linear action models.
\newblock In {\em Proceedings of the AAAI Conference on Artificial
  Intelligence}, volume~26.

\bibitem[Yao et~al., 2014]{yao2014pseudo}
Yao, H., Szepesv{\'a}ri, C., Pires, B.~A., and Zhang, X. (2014).
\newblock Pseudo-mdps and factored linear action models.
\newblock In {\em 2014 IEEE Symposium on Adaptive Dynamic Programming and
  Reinforcement Learning (ADPRL)}, pages 1--9. IEEE.

\bibitem[Yeong et~al., 2021]{yeong2021sensor}
Yeong, D.~J., Velasco-Hernandez, G., Barry, J., Walsh, J., et~al. (2021).
\newblock Sensor and sensor fusion technology in autonomous vehicles: A review.
\newblock {\em Sensors}, 21(6):2140.

\bibitem[Yurtsever et~al., 2020]{yurtsever2020survey}
Yurtsever, E., Lambert, J., Carballo, A., and Takeda, K. (2020).
\newblock A survey of autonomous driving: Common practices and emerging
  technologies.
\newblock {\em IEEE Access}, 8:58443--58469.

\bibitem[Zablocki et~al., 2022]{zablocki2022explainability}
Zablocki, {\'E}., Ben-Younes, H., P{\'e}rez, P., and Cord, M. (2022).
\newblock Explainability of deep vision-based autonomous driving systems:
  Review and challenges.
\newblock {\em International Journal of Computer Vision}, pages 1--28.

\bibitem[Zhang et~al., 2018]{zhang2019quota}
Zhang, S., Mavrin, B., Kong, L., Liu, B., and Yao, H. (2018).
\newblock {QUOTA:} the quantile option architecture for reinforcement learning.
\newblock {\em CoRR}, abs/1811.02073.

\bibitem[Zhou et~al., 2016]{zhou2016learning}
Zhou, B., Khosla, A., Lapedriza, A., Oliva, A., and Torralba, A. (2016).
\newblock Learning deep features for discriminative localization.
\newblock In {\em Proceedings of the IEEE Conference on Computer Vision and
  Pattern Recognition}, pages 2921--2929.

\end{thebibliography}

\end{document}